# Machine Learning for Multimodal Electronic Health Records-based Research: Challenges and Perspectives

Ziyi Liu[1,2], Jiaqi Zhang[1], Yongshuai Hou[1], Xinran Zhang[1,3], Ge Li, Yang Xiang[1*]

[1]Department of Network Intelligence, Peng Cheng Laboratory, Shenzhen, China

[2]School of Computer Science, Carnegie Mellon University, Pittsburgh, USA

[3]Mathematical and Computational Sciences, University of Toronto Mississauga, Mississauga, Canada

[4]School of Electronic and Computer Engineering, Peking University, Shenzhen, China

*Corresponding author

## Abstract

### Background

Electronic Health Records (EHRs) contain rich information of patients' health history, which usually include both structured and unstructured data. There have been many studies focusing on distilling valuable information from structured data, such as disease codes, laboratory test results, and treatments. However, relying on structured data only might be insufficient in reflecting patients' comprehensive information and such data may occasionally contain erroneous records.

### Objective

With the recent advances of machine learning (ML) and deep learning (DL) techniques, an increasing number of studies seek to obtain more accurate results by incorporating unstructured free-text data as well. This paper reviews studies that use multimodal data, i.e. a combination of structured and unstructured data, from EHRs as input for conventional ML or DL models to address the targeted tasks.

**Materials and Methods**

We searched in the Institute of Electrical and Electronics Engineers (IEEE) Digital Library, PubMed, and Association for Computing Machinery (ACM) Digital Library for articles related to ML-based multimodal EHR studies.

**Results and Discussion**

With the final 94 included studies, we focus on how data from different modalities were combined and interacted using conventional ML and DL techniques, and how these algorithms were applied in EHR-related tasks. Further, we investigate the advantages and limitations of these fusion methods and indicate future directions for ML-based multimodal EHR research.

**Introduction**

Electronic Health Records (EHRs) offer an efficient way to maintain patient information and are becoming more and more widely used by healthcare providers around the world. [1] Both structured and unstructured free-text data could be stored in EHRs in an organized way. Structured data can include numerical data such as lab test results, coded data such as procedure codes and diagnosis codes, categorical data such as medication lists and vital signs, and demographic information. Unstructured data such as clinical notes and discharge summaries can record more expressive and detailed descriptions of the patient's pain level, differences from the last visit, unique characteristics of their case, and anything that the physician thinks might be helpful for diagnosis and treatment, and supplement the nuances not seen in structured data.

[2] For example, Fox et al. validated a sample of hip fracture medical records and showed that coded records only have 66.7% sensitivity and 78.9% specificity for representing complications compared to postoperative notes. [3] By means of EHRs, physicians are able to keep track of patients' health status and visit history to facilitate diagnosis. Meanwhile, informaticians could develop algorithms to extract patterns from the vast amount of medical data of EHRs and provide insightful suggestions to physicians' decision-making. [4,5]

Machine learning (ML) have been used in a broad range of predictive tasks in healthcare, e.g. drug-drug interaction recognition, [6] disease progress prediction [7] and tumor detection. [8] In the past decades, traditional ML methods for EHR mining focused primarily on using structured data, which can usually be easily processed and analyzed using computer programs, and have demonstrated the effectiveness of corresponding predictive variables. [4,9] However, such structured data may suffer from missing- or erroneous-value problems, and thus cannot always guarantee consistent and accurate prediction results when used alone, as pointed out by [10,11]. Unstructured free-text data are harder to understand by programs, but are less prone to such errors, and it has been shown that textual features extracted by Natural Language Processing (NLP) pipelines can lead to positive results and serve as compensation to structured features. [12] With the development of more advanced ML and deep learning (DL) techniques, studies started to merge the two modalities, which are complementary in nature, to fully exploit the predictive potential of EHRs. Modality fusion strategies play a significant role in these studies.

There have been many reviews summarizing studies with EHR data as shown in Table 1. However, most reviews either do not discuss the fusion strategies of modalities [13–15] or focus on conventional ML models only. [12,16,17] [18] discusses fusion strategies in DL, but is of structured data and imaging data. In addition, some reviews are limited to one type of disease [12] or one type of task. [12,15,16] In our paper, we focus exclusively on studies that use conventional ML and DL techniques with multimodal EHR data, and place our emphasis on the strategies for combining different modalities. This definition of multimodality within this paper refers to structured data and unstructured free-texts in EHRs, which is converging with heterogeneity, but different from the traditional definition that often also includes audio, image, and videos as data sources. We don't place restrictions on certain types of diseases or tasks.

In the following sections, we review the cohort that supports multimodal analysis, tasks that involve multimodal analysis, and ML models and fusion strategies used in these studies. We also introduce a new taxonomy, interaction strategies, that is targeted to aligning the simple-form structured data to different processing granularities of unstructured data beyond fusion strategies. We thus provide a comprehensive analysis of recent progress in multimodal EHR analysis. In addition, we summarize the limitations of these methodologies and present potential future directions in this field. We expect that through this review, researchers can have a thorough view of the advancement of ML techniques in combing multimodal EHRs and a better understanding of how ML and DL models could be designed to align data from different modalities.

Table 1. Existing reviews with similar studied aspects.

|  | **Modalities of Included Studies** | **ML/DL** | **Main Focus and Whether Modality Fusion Discussed** | **Task** |
| --- | --- | --- | --- | --- |

| Shickel, 2018 [13] | Primarily structured + a few with both structured and unstructured | DL | DL techniques and their application to different EHR-based clinical tasks. Presents modality fusion as future direction. | All clinical tasks regardless of application domain |
| --- | --- | --- | --- | --- |
| Xiao, 2018 [14] | Primarily structured + a few with both structured and unstructured | DL | Identifying analytical tasks and introduces commonly used models for each task. Presents modality fusion as future direction. | All clinical tasks regardless of application domain |
| Huang, 2020 [18] | Imaging + structured data | DL | Fusion strategies of the modalities and implementation guidelines of multimodal models. | All clinical tasks regardless of application domain |
| Si, 2021 [15] | Multimodal + single-modal | DL | Resources, methods, applications and potential of EHR representation learning. Presents modality fusion as future direction. | All representation learning studies regardless of application domain |

| | Clinical notes. | ML + three DL articles | Application of NLP to clinical notes in 10 different chronic diseases groups. Does not discuss modality fusion. | Chronic diseases |
|---|---|---|---|---|
| Sheikhalishahi, 2019 [17] | | | | |
| Zeng, 2019 [16] | Primarily structured + a few with both structured and unstructured. | Rule-based, ML, DL | Application and state-of-the-art of NLP methods for computational phenotyping. Presents modality fusion as future direction. | Six tasks of computational phenotyping regardless of application domain |
| Ford, 2016 [12] | Primarily structured and unstructured + a few with unstructured only. | Rule-based, ML | Information extraction methods from unstructured data and improvement over structured data only. Does not focus on modality fusion. | Case-detection for named clinical conditions |

**Search Methods**

The focus of this review is on the application of conventional ML and DL techniques to multimodal EHR-based tasks, in which multimodal is defined as including both structured and unstructured free-text data. To obtain the set of related articles, we first searched in Institute of

Electrical and Electronics Engineers (IEEE) Digital Library, PubMed, and Association for Computing Machinery (ACM) Digital Library with query terms in the following format: ("multimodal" AND ("machine learning" OR "deep learning") AND "EHR" NOT "imaging"). We used various forms of each keyword, and included different ML and DL algorithms. The search resulted in 598 studies. Our inclusion criteria are that the research study must be using both structured and unstructured free-text EHR data in the proposed model, and that the model must use conventional ML or DL techniques. Based on these criteria, we screened 222 articles according to the title and abstract at the first step, and after full-text review, 82 articles were left. We then snowballed on other relevant review articles and included studies. Our final pool has 94 studies.

During full text review, we analyzed the studies from three perspectives: *data*, *task*, and *model*. For data, we focused on the dataset used by the studies, including size, language, availability and data types used (categorical, numerical, free-text, etc.). For task, we looked at where and how the proposed model is applied, and recorded elements such as task type, disease domain, and outcome. Finally, for the model, we focused on how each modality is represented, categorized the fusion strategy of the modalities, and recorded implementation details of the model.

**Cohorts**

The population size in each study varies greatly from hundreds to millions (of patients). For example, [19] used a dataset with only 300 patients to train a complex disease identification model, while [20] used a dataset of 6 million patients for Atopic Dermatitis identification. The dataset size used in most studies is over 1,000. Cohorts from large datasets are mostly generated automatically according to certain criteria (i.e. phenotyping) related to the research aims [21], while the cohorts from most small datasets were often manually chosen and annotated, which could have higher accuracy and be more targeted towards their tasks. [22] Moreover, most large datasets are in English with patients from the US and UK, and datasets in other languages such as Chinese, [23] Dutch, [24] Swedish, [25,26] and Japanese [26] tend to be smaller.

The multimodal EHR datasets generally contain clinical text (e.g. clinical notes), codes (e.g. ICD codes), categorical data (e.g. medication list), and numerical data (e.g. laboratory measurements). Most datasets also include multi-visit information, although some proposed models did not specifically model such sequential information. [20,22,27] Most studies used private datasets, while others used publicly available datasets [28–30] or shared their datasets. [21,31] The most popular public dataset is the Medical Information Mart for Intensive Care III (MIMIC-III), which comprises de-identified health-related data associated with 53,423 distinct hospital admissions for adult patients. [32] Other public datasets such as Vanderbilt University Medical Center EHR, [33] Mount Sinai Data Warehouse [34] and I2B2 [35] were also used more than once in the included studies and can be used to conduct further multimodal EHR research.

**Tasks**

Traditional ML-based tasks related to EHRs can be roughly divided into *clinical information regularization* and *clinical decision-making*. Usually, unstructured information, such as case reports and nursing notes, is used as the main data source in the first type of task. The information is extracted through NLP processes and transformed into a structured format that is easier to store and analyze. [36] In comparison, for the second task, structured information such as ICD codes and patient demographics data is more often used. [37,38] EHR analysts have also been exploring the fusion of the two types of information. For example, Payrovnaziri and Barrett extended their prior work by adding unstructured text features into previously used structured data to predict one-year mortality risk of patients with AMI. [39] Experiments demonstrate that the performance of the algorithmic model that combines both structured and unstructured information is superior to the results obtained when only one of them is used. [40]

According to our review, multimodal data that combines structured and unstructured information is currently mainly used in the clinical decision-making task, such as the prediction of different types of diseases and patient risk assessment. [41–44] Another direction, which is more related to the clinical information regularization task is patient representation learning, where researchers attempt to model multimodal EHR data in a shared semantic space for general clinical tasks. [28] The performance of these learned representations is verified on baseline tasks such as the prediction of hospital length of stay and the rate of readmission, and shows that representation learned from multimodal data has a more comprehensive generalization ability in solving real-world problems. [7] Other tasks with multimodal EHR data, such as patient trail matching and improving phenotyping algorithms, have also presented promising results. [45] These studies may provide further directions on applicable scenarios of multimodal data for EHR researchers, and the improved performance on a wide range of tasks has demonstrated the properness and explorable space of this direction.

**Machine Learning Methods**

In total, there are 63 (67.0%) papers using primarily conventional ML and 30 (31.9%) papers using primarily DL, with one paper not explicitly introducing methods or models used but mentioning they used ML. As discussed above, most tasks with multimodal EHR deal with classification (clinical decision-making) or patient representation problems. For classification tasks such as disease prediction and risk assessment, relevant studies used mainly conventional ML models, such as Logistic Regression (LR), [20,46,47] Random Forest (RF), [21,46,48] Support Vector Machines (SVMs), [29,49,50] and Naive Bayes (NB). [27,49,50] For instance, Chen at el. took the UMLS concepts extracted from clinical notes and billing codes as input and incorporated active learning to SVM-based phenotyping algorithms. [51] Slightly different from traditional approaches, Zhao and Weng designed a weighted Bayesian Network Inference (BNI) model where they combined structured EHR data with a prior probability calculated from free-text PubMed abstracts of the expert selected variables for pancreatic cancer prediction. [52]

Despite the popularity of conventional ML for classification tasks, DL models were greatly explored over the recent years, and are being applied to both classification tasks and representation learning tasks. [53–55] CNN has been shown to be good at extracting locality-invariant features, which can be useful for identifying key concepts in clinical notes. [56] On the other hand, RNN, and especially its variant Long Short-Term Memory Networks (LSTM), [57] is well known for discovering sequential dependencies, which is often used to model clinical notes (text as sequence) and temporal information in EHR. [28] More recently, Transformer-based pre-trained models are also adapted to the medical domain. Huang et al. developed ClinicalBERT, which pre-trained the BERT model on clinical notes in MIMIC-III. [58] This model has also been used by many multimodal EHR studies to build embeddings for free-text data. [43,45,59] Among the included studies, Autoencoders and word embeddings are the most commonly used techniques to compute the representation. [34] Other approaches that applied recent DL techniques also exist. Lee et al. built a harmonized representation space for patients, medical concepts, and medical events by modeling the time-variant patients nodes with LSTM, and time-invariant concept and event nodes with Graph Convolutional Networks (GCN), which fused multimodal inputs. [60] After obtaining these representations, they can be used as input to both conventional ML and DL models for downstream tasks.

To compare, most conventional ML methods take each feature, either from structured or unstructured EHRs, independently owing to their shallow structures. For example, it is hard for RF and SVM to capture the sequential dependencies conveyed in unstructured data, and thus some deeper semantic information might be missed during modeling. On the other hand, DL models have a stronger capacity to model different levels of dependencies between feature dimensions so that complex interactions between modalities might be built. Specifically, DL models are superior in modeling free-text, e.g. using hierarchical and sequential structures, which is more conformable with the natural structures of free-text.

**Fusion Strategies**

The remarkable characteristic of multimodal EHR research is that useful information can be conveyed by both structured and unstructured data, and thus features extracted from the two modalities should be fused effectively. As a reference, there exists a generally accepted taxonomy by previous studies which categorizes multimodal data fusion in ML into three types: early fusion, joint fusion, and late fusion, as shown in Figure 1. [18] We firstly tried to align each study to this taxonomy, shown in Table 2.

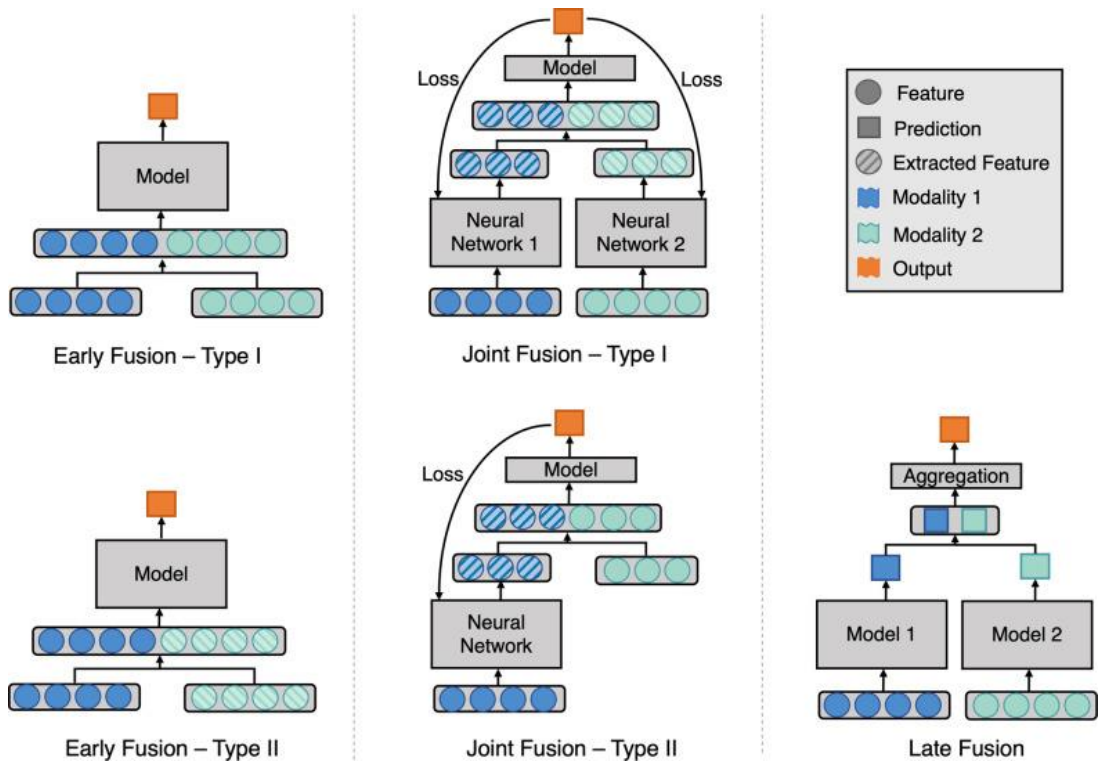

Figure 1. Early/Joint/Late Fusion by Huang et al., 2020 [18]

Table 2. Different fusion types for multimodal EHR data that can be aligned to.

| Fusion Type | Definition | Commonly Used By | Examples |
|---|---|---|---|
| Early Fusion | Features are extracted from modalities by statistical methods, existing NLP tools, word embedding models, or other DL models, and combined prior to passing to the model | ML, DL | statistical methods [26,48,61] Existing NLP tools [21,46,62] Word embedding models [39,63,64] Other DL models [22,59,65] |

| Joint Fusion | Combines the modalities at intermediary layers of the neural network, and losses can be propagated back to the feature extraction phase to dynamically update the feature weights | DL | [31,66–68] |
|---|---|---|---|
| Late Fusion | Separate models are used for each modality, and the final decision leverages the decision of each individual model by some ensemble strategies | ML | [48,63,69,70] |

**Interaction Strategies**

Unlike imaging and speech data which usually need to be preprocessed separately and transformed into feature vectors before fusion, free-text data is more flexible in the sense that its fusion with structured data can start at a much earlier stage and be in a more direct way because of the similarity in modality. Each word in free-text is semantically meaningful by itself, whereas extracting such segments from an image or a speech signal sequence often requires additional handling. Different levels of text processing could lead to features capturing different granularities of semantics. For example, raw free-text data could be used directly to compute word embeddings that can reflect relationships between words, and concepts and topics could be extracted from free-text to represent its meaning at the sentence/document level.

Therefore, as free-text data has higher granularities, the three fusion strategies alone might be insufficient to fully represent the various levels of interactions between multi modalities. For example, using raw words, word embeddings or named entities as input to a neural network model can all be aligned to early fusion, but they represent quite distinct combination scenarios where different complementary and interdependent information could be carried and different model structures could be applied. We design a novel and finer-grained taxonomy, *interaction strategies*, to better represent how multimodal data are fused and help us understand how different models deal with multimodal data according to their operational mechanism. We define five categories based on the degree to which free-text data is processed before combining with structured data (Figure 2). We also discuss pros and cons of each level of interaction in Table 3.

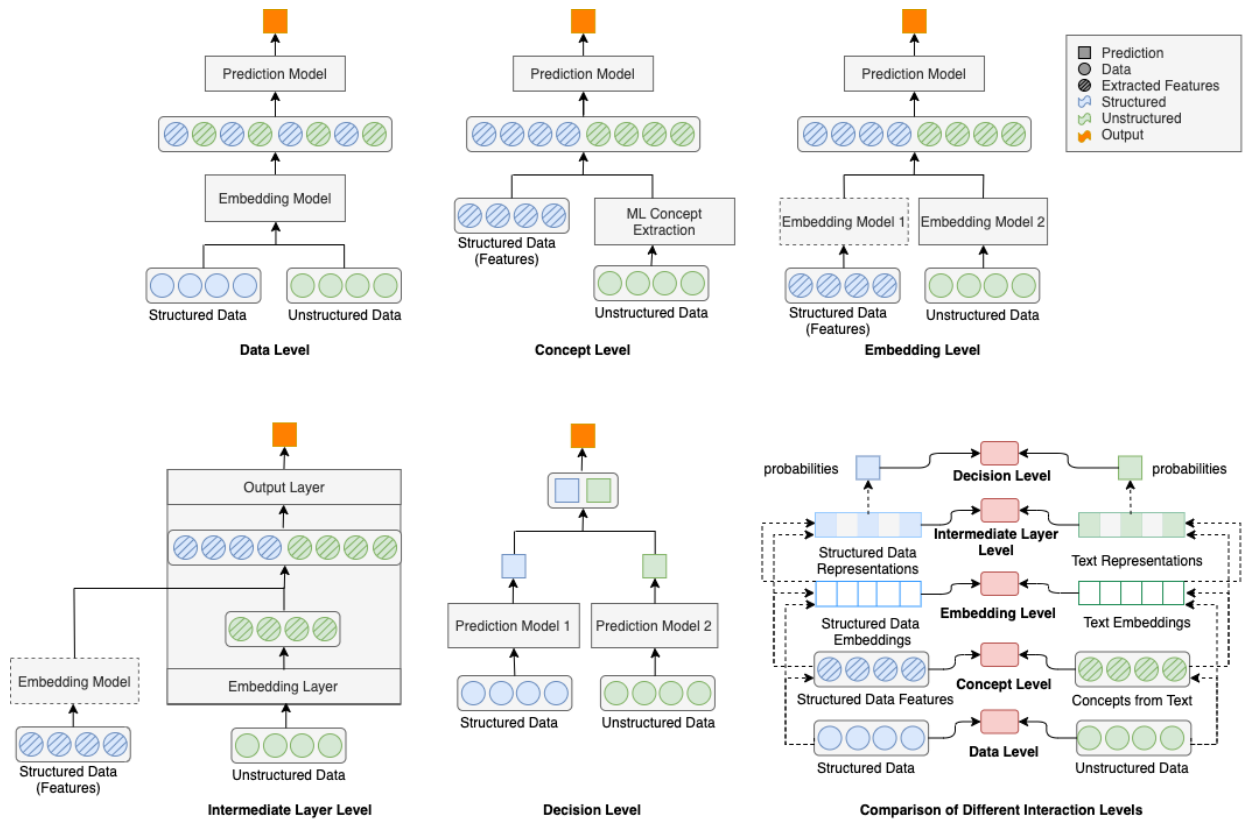

Figure 2. Different interaction levels for multidata fusion. We also demonstrate the comparison of different interaction levels in the last sub-figure. The legend is similar to that in Figure 1.

Table 3. Pros and cons of the five interaction strategies.

| Interaction Strategy | Pros | Cons |
|---|---|---|
| Data-level | Could lead to the discovery of mutual and representative features which are present in more than one modality. | Data from the two modalities are considered as coming from only one modality, e.g. consider a code as a word.<br><br>Resulting representation is more likely to capture shared rather than complementary information from the modalities. |
| Concept-level | Can be easily implemented and is computationally efficient.<br><br>Extracted textual features provide a clean and concise representation of the unstructured data, and thus increases the model's interpretability and is | Could lead to information loss, such as contextual and temporal information in free-text data.<br><br>Concatenation is a relatively naive way to combine modalities that does not exploit full relationships between the modalities. |
| | useful when performing feature analysis and ablation study. | |
| Embedding-level | Word embeddings and neural networks can greatly preserve information in free-text and well-formed patterns and features in structured data could be discovered than using frequency counts.<br><br>Dependency on external tools is reduced. | The boundary between modalities is blurred with concatenated single-modal representations, which might cause the model to rely on dominating modality instead of learning the cross-modality interaction.<br><br>Decreased model interpretability. |
| Intermediate Layer-level | Retains all advantages of embedding-level interaction and provides greater interaction flexibility.<br><br>Easier to learn complementary information from all modalities and dynamically update feature weights accordingly by incorporating the combination step within the model. | Low quality features from one modality could be harmful and add additional noise to the model.<br><br>It is debatable whether one or two layers could fully capture the complex relationships between the modalities depending on at which layer the interaction happens. |
| Decision-level | Performance of each modality can be evaluated separately, which can be useful when the number of modalities | Models in this interaction strategy are not exposed to multimodal data at all, and thus may not be able to |

| | is large, or there are multichannel representations, and decisions need to be made regarding which ones to keep. | effectively use the rich information contained in the multimodal data. |

Data-level Interaction

In data-level interaction, free-text data remains unprocessed and is combined with structured data in its raw form. Methods using this strategy usually take as input both structured and unstructured raw data and learn a shared embedding space by word embedding models. Bai et al. proposed JointSkip-gram, which is based on the Word2Vec learning schema. [71] It takes raw clinical notes and medical codes as input, and for each code, learns to predict all other codes and words in the same visit, while for each word, learns to predict its neighboring words and all codes. The resulting representation is able to capture not only similarities within the same modality, but also cross different modalities through shared weights.

Concept-level Interaction

Concept-level interaction refers to extracting features firstly by statistics or established packages from each modality, and combining them using simple ways such as concatenation. As the free-text features can be either raw data or clinical concepts, e.g. named entities, we call this interaction concept-level. This is a typical type of early fusion, and appears most often in conventional ML studies, before the advance of DL. Feature values of structured data are often represented as the frequencies of events, while for free-text data, they are extracted using Term Frequency-Inverse Document Frequency (TF-IDF), bag of words, topic modeling methods such as Latent Dirichlet Allocation (LDA), [72] or established NLP pipelines such as clinical Text

Analysis and Knowledge Extraction System (cTAKES), [73] MetaMap, [74] KnowledgeMap, [75] and Health Information Text Extraction (HITEx). [76] [62] is a typical example of concept-level interaction with machine learning, where features such as disease concepts, lab results, and medications were extracted from both codified structured data and clinical notes using HITEx. Then the frequencies of events were passed to penalized logistic regression to predict rheumatoid arthritis.

Embedding-level Interaction

Embedding-level interaction shows another way of integrating free-text, in which representation of free-text data is obtained by passing the raw text through a pre-trained word embedding or neural network model. Structured data can be represented similarly as in concept-level interaction or also be passed through a neural network model as embedding-like representations. The embedding vectors from each modality are then concatenated or summed up as the input for the final output model. This type of interaction is used by many DL studies according to our review due to the ability of embeddings to capture more comprehensive information from the raw data. Beeksma et al. embedded unstructured free-text data using Word2Vec, and concatenated with structured codes, laboratory, and medication data for life expectancy prediction using LSTM. [64] Darabi et al. developed two separate modules for code and text representation. [59] Code module is composed of a skip-gram model followed by a Transformer encoder [77] and the text model uses BioBERT followed by a Bi-GRU. Finally, patient representation is a concatenation of the code and text representations.

Intermediate Layer-level Interaction

In intermediate layer-level interaction, the interaction of modalities happens between the embedding and the output layer. Intermediate layer-level interaction is used only by DL studies, but less popular than embedding-level interaction. Liu et al. proposed CNN and Bi-LSTM based models for chronic disease prediction, which first represented free-text data using word embedding pre-trained on medical data, and passed to the base model. [31] Structured data were concatenated to the layer before the last fully connected layer. Xu et al. adopted a novel approach with Memory Networks, where they used clinical notes encoded by hierarchical LSTM as the query, and fed structured clinical sequences into the memory. [42] Then the information similar to the query (unstructured data) is extracted from the memory (structured data) and combined with the query to form a representation for acute kidney injury predictions.

Decision-level Interaction

Decision-level interaction is the same as late fusion, in which separate models are built for each modality, and modalities only interact with each other after the output layer. Shin et al. used TF-IDF and topic modeling to represent two types of clinical notes, and built separate logistic regression models for each feature set. [69] The final division is the average of all model outputs. This level of interaction is not commonly used in multimodal EHR studies, especially those with DL.

In addition to these semantic granularity-based interactions, there also are interaction scenarios that are specific to EHRs and from other perspectives. For example, the multimodal data can be fused in either the visit level, i.e. the interaction happens within each visit, or the patient level, i.e. the interaction happens after all the visits are encoded. In these scenarios, either of the above-mentioned interactions on different levels could be leveraged.

**Limitations**

The acquisition of multimodal EHR data is one of the most significant limitations. Since different EHR systems have different standards, not all systems store data as having both "structured" and "unstructured" parts, especially in different countries, where data might be all structured, all semi-structured, or all unstructured. This makes multimodal EHR datasets more scarce than single-modal data. And as mentioned in the cohort section, most current multimodal EHR data are in English, it could introduce distribution biases on population that might lead to algorithmic biases. Finally, unlike structured data, unstructured data is more sensitive to the missing data problem. It is possible to impute a missing blood pressure value, but difficult to impute a missing clinical note.

Another limitation is on the modality interaction techniques and models used. Although the advance of DL propels new techniques such as embedding and intermediate layer level interaction, the "interaction" strategy in these studies is still simple and possibly insufficient to capture complex interactions between different modalities. In addition, base models in these studies are often restricted to traditional ML and simple DL models This could be due to the fact that processing and extracting useful information from free-text data by traditional informaticians, statisticians or physicians remains a challenge. There have been studies using more complex models such as Graph Convolutional Networks to wrap up data from multimodalities,[60] but the number remains limited, and the exploration is still incipient.

**Future Directions**

*Include more data modalities*

Although we focus on structured and unstructured text as our multimodal data, other modalities such as images and videos could also be integrated using the aforementioned fusion or interaction techniques. [18,78] Integrating more data modalities could be helpful if each data modality contains incomplete but complementary information, which is frequently seen in EHRs. For example, Patient history, imaging diagnostic report and associated lab test values could naturally be used together with medical imaging for making clinical decisions. [78] However, whether to include data from more sources is also restricted by the data availability and sharing policies in many cases. Furthermore, in some scenarios, prior knowledge, including terminologies, ontologies and knowledge bases, have also been proven useful when performing ML-aided clinical prediction. [52,68] It can serve as another data modality and the relevant techniques deserve a deeper investigation.

*Explore more ways of interaction between modalities*

There are still many limitations in the fusion methods research of multimodal data. This issue is also a major direction that needs attention in future research. Most of the current research uses methods of direct concatenation to process multimodal data, either from data or embedding level. This method is easy to implement, but the representation gap of different types of data is ignored during this process. For example, embeddings trained from unstructured data are quite different from those trained from structured data in the semantic level, e.g. the semantic granularity is different. [79] Future research on the fusion strategy of multimodal data should find more ways to align the representation level of different types of data. Current works have tried different solutions, but a consensus on a widely accepted strategy for multimodal interaction, especially in the fusion method of structured and unstructured data, has not been retained yet. The interaction strategies of multimodal methods using different types of data, such as interaction of image data [80], structured data and text data, may be worth exploration and adoption among each other.

*How and how much pretrained models can help?*

Pretrained models (PTM) have been proved helpful in many natural language processing [81] as well as multimodal classification tasks (e.g. integrating image and text). [82] Using PTMs, comprehensive contextual information for both different modalities and their interactions could be learned in a self-supervised way [83] or with minimum supervisions [84]. For EHR mining, however, the explorations on multimodal data that contain both structured and unstructured text are limited. [71] is one representative study in this paradigm but it only applies shallow embedding techniques and fails to encoding deeper contextual information. [85] is a pioneer study that pretrains structured and unstructured data together using BERT, but it only uses the diagnosis information as the structured modality. Adopting more techniques from the general multimodal pretraining (e.g. VideoBERT [86]) or treating it as a machine translation problem [87] might be ways for further exploration.

*Enable more flexible data acquisition and sharing*

As mentioned above, the availability of multimodal data is one of the limitations that hinder the development of corresponding research. There are many factors (e.g. gender, ethnicity, political issues, weather, region, environmental humidity) that could influence the research directions or even clinical decisions, it might not be sufficient to rely on the few public datasets to make conclusive clinical claims to the general population. Therefore, it is necessary to encourage more institutions or hospitals to share flexible data to include a more general and broader range of population to facilitate clinical research. In ML, federated learning (FL) [88] has the capability to collect patients' data with the safety of patients' privacy protection and data security from multicenter. It might be leveraged to collect multimodal EHR features from multicenters to train a large-scale model in a discrete distribution without collecting patients' EHR directly.

**Conclusion**

This review summarizes current advances in multimodal EHR studies using both structured and unstructured free-text EHR data with conventional ML and DL techniques. We proposed interaction strategies, a new taxonomy targeted to the combination of structured and free-text EHR data. Our finding suggests that there is a growing interest in multimodal EHR, but most studies combine the modalities with relatively simple strategies, which despite being shown to be effective, might not fully exploit the rich information embedded in these modalities. We acknowledge that there are still limitations with this review such as the limitation in the coverage of queried papers and included algorithms. As this is a fast-growing field and new models are constantly being developed, there might exist studies that fall outside of our definition of strategies or use a combination of these strategies. Nonetheless, we believe that the development of this field will give rise to more comprehensive EHR analysis and will be of great support to the clinical decision-making process.

**Funding**

There is no funding resource to be disclosed.

## Author contributions

YX and ZYL conceived the study. ZYL designed search method, analyzed model aspect of included papers, and drafted most parts of the manuscript. JQZ participated in study design, analyzed task aspect of included papers, and drafted Task section. YSH analyzed cohort aspect of included papers, and drafted Cohort section. XRZ contributed to future direction, YX supervised the research, provided feedback to the proposed taxonomy, contributed to future direction, and critically revised the manuscript. GL participated in manuscript review. All authors provided feedback and approved the final version of the manuscript.

## Conflict of interest statement